\documentclass{article}

\PassOptionsToPackage{numbers, round,longnamesfirst, compress, square, sort}{natbib}
\usepackage[final]{nips_2016}

\usepackage[utf8]{inputenc} 
\usepackage[T1]{fontenc}    
\usepackage{hyperref}       
\usepackage{url}            
\usepackage{booktabs}       
\usepackage{amsfonts}       
\usepackage{nicefrac}       
\usepackage{microtype}      
\usepackage{bbm}
\usepackage{algorithm}
\usepackage{amssymb}
\usepackage{amsmath}
\usepackage{epsfig}
\usepackage{graphicx}
\usepackage{amsmath}
\usepackage{url} 
\usepackage{amssymb}
\usepackage{multirow}
\usepackage{graphicx}
\usepackage{fancyvrb}
\usepackage{color}
\usepackage{amsmath}
\usepackage{graphicx}
\usepackage{caption}
\usepackage{bbm}
\usepackage{algpseudocode}
\usepackage{float}
\usepackage{newverbs}
\usepackage{microtype}
\usepackage{xcolor}
\usepackage{breqn}
\usepackage{wrapfig}
\usepackage{soul}
\usepackage[colorinlistoftodos]{todonotes}
\usepackage[subrefformat=parens,labelformat=parens]{subfig}

\title{Deep Successor Reinforcement Learning}

\begin{document}

\author{Tejas D. Kulkarni\thanks{Authors contributed equally and listed alphabetically.} \\
		BCS, MIT \\
        \texttt{tejask@mit.edu}  \And
        Ardavan Saeedi$^*$ \\
        CSAIL, MIT \\
        \texttt{ardavans@mit.edu} \And
        Simanta Gautam \\ 
        CSAIL, MIT \\
        \texttt{simanta@mit.edu} \And
        Samuel J. Gershman\\
		Department of Psychology\\
	    Harvard University\\
	    {\tt gershman@fas.harvard.edu}
}

\maketitle

\begin{abstract}
Learning robust value functions given raw observations and rewards is now possible with model-free and model-based deep reinforcement learning algorithms. There is a third alternative, called Successor Representations (SR), which decomposes the value function into two components -- a reward predictor and a successor map. The successor map represents the expected future state occupancy from any given state and the reward predictor maps states to scalar rewards. The value function of a state can be computed as the inner product between the successor map and the reward weights. In this paper, we present DSR, which generalizes SR within an end-to-end deep reinforcement learning framework. DSR has several appealing properties including: increased sensitivity to distal reward changes due to factorization of reward and world dynamics, and the ability to extract bottleneck states (subgoals) given successor maps trained under a random policy. We show the efficacy of our approach on two diverse environments given raw pixel observations -- simple grid-world domains (MazeBase) and the \textit{Doom} game engine. \footnote{Code and other resources -- \url{https://github.com/Ardavans/DSR} }

\end{abstract}

\section{Introduction}
Many learning problems involve inferring properties of temporally extended sequences given an objective function. For instance, in reinforcement learning (RL), the task is to find a policy that maximizes expected future discounted rewards (value). RL algorithms fall into two main classes: (1) model-free algorithms that learn cached value functions directly from sample trajectories, and (2) model-based algorithms that estimate transition and reward functions, from which values can be computed using tree-search or dynamic programming. However, there is a third class, based on the \emph{successor representation} (SR), that factors the value function into a predictive representation and a reward function. Specifically, the value function at a state can be expressed as the dot product between the vector of expected discounted future state occupancies and the immediate reward in each of those successor states.

Representing the value function using the SR has several appealing properties. It combines computational efficiency comparable to model-free algorithms with some of the flexibility of model-based algorithms. In particular, the SR can adapt quickly to changes in distal reward, unlike model-free algorithms. In this paper, we also highlight a feature of the SR that has been less well-investigated: the ability to extract bottleneck states (candidate subgoals) from the successor representation under a random policy \cite{stachenfeld2014design}. These subgoals can then be used within a hierarchical RL framework. In this paper we develop a powerful function approximation algorithm and architecture for the SR using a deep neural network, which we call \emph{Deep Successor Reinforcement Learning} (DSR). This enables learning the SR and reward function from raw sensory observations with end-to-end training.

\begin{figure*}
  \centering
  \includegraphics[width=4.7in]{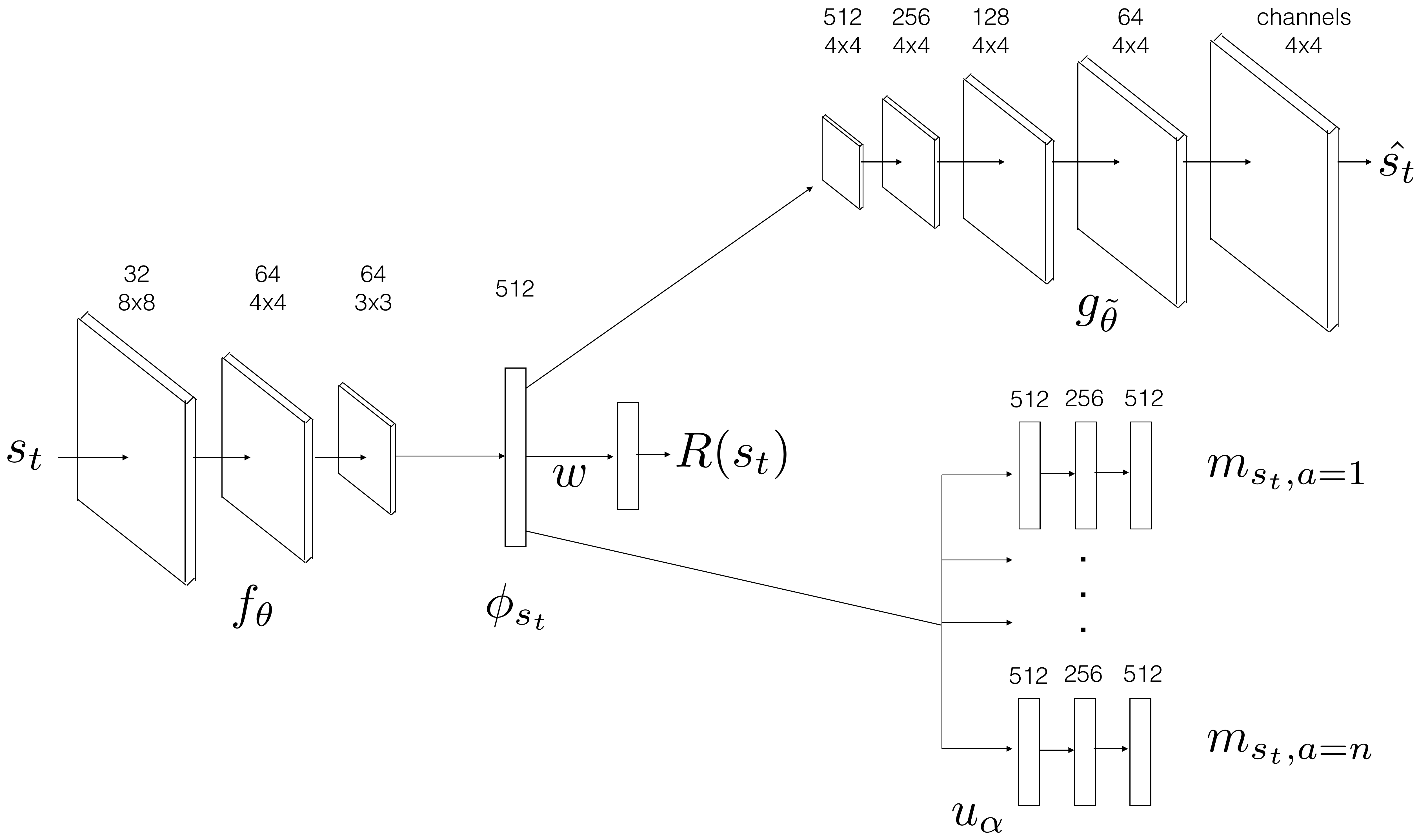}
  \caption{\textbf{Model Architecture:} DSR consists of: (1) feature branch $f_{\theta}$ (CNN) which takes in raw images and computes the features $\phi_{s_t}$, (2) successor branch $u_{\alpha}$ which computes the SR $m_{{s_t},a}$ for each possible action $a \in \mathcal{A}$, (3) a deep convolutional decoder which produces the input reconstruction $\hat{s_t}$ and (4) a linear regressor to predict instantaneous rewards at $s_t$. The Q-value function can be estimated by taking the inner-product of the SR with reward weights: $Q^{\pi}(s,a)\approx m_{sa} \cdot \mathbf{w}$.}.
  \label{fig:arch}
\end{figure*}

The DSR consists of two sub-components: (1) a reward feature learning component, constructed as a deep neural network, predicts intrinsic and extrinsic rewards to learn useful features from raw observations; and (2) an SR component,  constructed as a separate deep neural network, that estimates the expected future ``feature occupancy'' conditioned on the current state and averaged over all actions. The value function can then be estimated as the dot product between these two factored representations. We train DSR by sampling experience trajectories (state, next-state, action and reward) from an experience replay memory and apply stochastic gradient descent to optimize model parameters. To avoid instability in the learning algorithm, we interleave training of the successor and reward components. 

We show the efficacy of our approach on two different domains: (1) learning to solve goals in grid-world domains using the \textit{MazeBase} game engine and (2) learning to navigate a 3D maze to gather a resource using the \textit{Doom} game engine. We show the empirical convergence results on several policy learning problems as well as sensitivity of the value estimator given distal reward changes. We also demonstrate the possibility of extracting plausible subgoals for hierarchical RL by performing normalized-cuts on the SR~\cite{shi2000normalized}.

\section{Related work}
The SR has been used in neuroscience as a model for describing different cognitive phenomena. \cite{gershman2012successor} showed that the temporal context model \cite{howard2002distributed}, a model of episodic memory, is in fact estimating the SR using the temporal difference algorithm.  \cite{corneil2015attractor} introduced a model based on SR for preplay and rapid path planning in the CA3 region of the hippocampus. They interpret the SR as an an attractor network in a low–dimensional space and show that if the network is stimulated with a goal location it can generate a path to the goal. \cite{stachenfeld2014design} suggested a model for tying the problems of navigation and reward maximization in the brain. They claimed that the brain's spatial representations are designed to support the reward maximization problem (RL); they showed the behavior of the place cells and grid cells can be explained by finding the optimal spatial representation that can support RL. Based on their model they proposed a way for identifying reasonable subgoals from the spectral features of the SR. Other work (see for instance, \cite{botvinick2014model, daw2014algorithmic}) have also discussed utilizing the SR for subgoal and option discovery.

There are also models similar to the SR that have been been applied to other RL-related domains. \cite{schraudolph1994temporal} introduced a model for evaluating the positions in the game of Go; the model is reminiscent of SR as it predicts the fate of every position of the board instead of the overall game score. Another reward-independent model, universal option model (UOM), proposed in \cite{szepesvari2014universal}, uses state occupancy function to build a general model of options. They proved that UOM of an option, given a reward function, can construct a traditional option model. There has also been a lot of work on option discovery in the tabular setting \cite{mcgovern2001automatic, csimcsek2005identifying, mannor2004dynamic, menache2002q, konidaris2009skill}. In more recent work, Machado et al. \cite{machado2016learning} presented an option discovery algorithm where the agent is encouraged to explore regions that were previously out of reach. However, option discovery where non-linear state approximations are required is still an open problem. 

Our model is also related to the literature on value function approximation using deep neural networks. The deep-Q learning model \cite{mnih2015human} and its variants (e.g., \cite{silver2016mastering, schaul2015prioritized, nair2015massively, mnih2016asynchronous}) have been successful in learning Q-value functions from high-dimensional complex input states.

\section{Model}
\subsection{Background}
Consider an MDP with a set of states $\mathcal{S}$, set of actions $\mathcal{A}$, reward function $R: \mathcal{S} \rightarrow \mathbb{R}$, discount factor $\gamma \in [0, 1]$, and a transition distribution $T: \mathcal{S} \times \mathcal{A} \rightarrow [0,1]$. Given a policy $\pi : \mathcal{S} \times \mathcal{A} \rightarrow [0,1]$, the Q-value function for  selecting action $a$ in state $s$ is defined as the expected future discounted return:
\begin{align}
Q^{\pi}(s,a) = \mathbb{E}\left[\sum^{\infty}_{t=0}\gamma^tR(s_t)| s_0 = s, a_0 = a\right],
\end{align}
where, $s_t$ is the state visited at time $t$ and the expectation is with respect to the policy and transition distribution. The agent's goal is to find the optimal policy $Q^*$ which follows the \textit{Bellman equation}:
\begin{equation}
\label{eq:Q_bellman}
Q^*(s,a) = R(s_t) + \gamma \max_{a'} \mathbb{E}\left[Q(s_{t+1}, a')\right].
\end{equation}

\subsection{The successor representation}
The SR can be used for calculating the Q-value function as follows. Given a state $s$, action $a$ and future states $s'$, SR is defined as the expected discounted future state occupancy:
\[
M(s, s', a) =\mathbb{E}\left[\sum^{\infty}_{t=0}\gamma^t\mathbbm{1}[s_t=s'] |s_0=s, a_0=a\right] ,
\]
where $\mathbbm{1}[.]=1$ when its argument is true and zero otherwise. This implicitly captures the state visitation count. Similar to the Bellman equation for the Q-value function (Eq. \ref{eq:Q_bellman}), we can express the SR in a recursive form:
\begin{equation} 
\label{eq:sr_bellman}
M(s,s',a) = \mathbbm{1}[s_t = s'] + \gamma \mathbb{E}[M(s_{t+1},s',a_{t+1})].
\end{equation}
Given the SR, the Q-value for selecting action $a$ in state $s$ can be expressed as the inner product of the immediate reward and the SR \cite{dayan1993improving}:
\begin{equation} \label{eq:1}
Q^{\pi}(s,a)=\sum_{s'\in \mathcal{S}}M(s,s',a)R(s')
\end{equation}

\subsection{Deep successor representation}
For large state spaces, representing and learning the SR can become intractable; hence, we appeal to non-linear function approximation. We represent each state $s$ by a $D$-dimensional feature vector $\phi_s$ which is the output of a deep neural network $f_{\theta}:\mathcal{S}\rightarrow \mathbb{R}^D$ parameterized by $\theta$.

For a feature vector $\phi_s$, we define a feature-based SR as the expected future occupancy of the features and denote it by $m_{sa}$. We approximate $m_{sa}$ by another deep neural network $u_\alpha$ parameterized by ~$\alpha$: $m_{sa} \approx u_{\alpha}(\phi_s, a)$. We also approximate the immediate reward for state $s$ as a linear function of the feature vector $\phi_S$: $R(s) \approx \phi_s \cdot \mathbf{w}$, where $\mathbf{w} \in \mathbb{R}^D$ is a weight vector. Since reward values can be sparse, we can also train an intrinsic reward predictor $R_i(s) = g_{\tilde{\theta}}(\phi_s)$. A good intrinsic reward channel should give dense feedback signal and provide features that preserve latent factors of variations in the data (e.g. deep generative models that do reconstruction). Putting these two pieces together, the Q-value function can be approximated as (see \ref{eq:1} for closed form):
\begin{align}
Q^{\pi}(s,a)\approx m_{sa} \cdot \mathbf{w}.
\end{align}
The SR for the optimal policy in the non-linear function approximation case can then be obtained from the following Bellman equation:
\begin{align}
m_{sa} = \phi_s + \gamma \mathbb{E}\left[ m_{s_{t+1}a'}\right]
\end{align}
where $a' = \text{argmax}_a m_{s_{t+1}a} \cdot \textbf{w}$.

\subsection{Learning}

The parameters $(\theta, \alpha, \textbf{w}, \tilde{\theta})$ can be learned online through stochastic gradient descent. 

The loss function for $\alpha$ is given by:
\[
L^m_t(\alpha, \theta) = \mathbb{E}[(\phi(s_t) + \gamma u_{\alpha_{prev}}(\phi_{s_{t+1}}, a') - u_{\alpha}(\phi_{s_t},a))^2],
\]
where $a' = \text{argmax}_a u_{\alpha}(\phi_{s_{t+1}},a) \cdot \textbf{w}$ and the parameter $\alpha_{prev}$ denotes a previously cached parameter value, set periodically to $\alpha$. This is essential for stable Q-learning with function approximations (see \cite{mnih2015human}).

For learning $\mathbf{w}$, the weights for the reward approximation function, we use the following squared loss function:
\begin{equation}
\label{eq:r_loss}
L^r_{t}(\mathbf{w}, \theta) = (R(s_t) - \phi_{s_t} \cdot \mathbf{w})^2
\end{equation}

Parameter $\theta$ is used for obtaining the $\phi(s)$, the shared feature representation for both reward prediction and SR approximation. An ideal $\phi(s)$ should be: 1) a good predictor for the immediate reward for that state and 2) a good discriminator for the states. The first condition can be handled by minimizing loss function $L^r_{t}$; however, we also need a loss function to help in the second condition. To this end, we use a deep convolutional auto-encoder to reconstruct images under an L2 loss function. This dense feedback signal can be interpreted as an intrinsic reward function. The loss function can be stated as:
\begin{equation}
L^a_t(\tilde{\theta}, \theta)= (g_{\tilde{\theta}}(\phi_{s_t}) - s_t)^2.
\end{equation}
The composite loss function is the sum of the three loss functions given above: 
\begin{equation}
L_t(\theta, \alpha, \textbf{w}, \tilde{\theta}) =
L^m_t(\alpha, \theta) + 
L^r_t(\mathbf{w}, \theta) + L^a_t(\tilde{\theta}, \theta) 
\label{eq:comp_loss}
\end{equation}
For optimizing Eq. \ref{eq:comp_loss}, with respect to the parameters $(\theta, \alpha, \textbf{w}, \tilde{\theta})$, we iteratively update $\alpha$ and $(\theta, \textbf{w}, \tilde{\theta})$. That is, we learn a feature representation by minimizing $L^r_t(\mathbf{w}) + L^a_t(\tilde{\theta})$; then given $(\theta^*, \textbf{w}^*, \tilde{\theta}^*)$, we find the optimal $\alpha^*$. This iteration is important to ensure that the successor branch does not back-propagate gradients to affect $\theta$. We use experience replay memory $\mathcal{D}$ of size $1e^6$ to store transitions, and apply stochastic gradient descent with a learning rate of $2.5e^{-4}$, momentum of $0.95$, a discount factor of $0.99$ and the exploration parameter $\epsilon$ annealed from 1 to 0.1 as training progresses. Algorithm 1 highlights the learning algorithm in greater detail.

\begin{algorithm}
\caption{Learning algorithm for DSR}
\label{alg:training}
\begin{algorithmic}[1]
\State Initialize experience replay memory $\mathcal{D}$, parameters $\{\theta,\alpha, \mathbf{w}, \tilde{\theta}\}$ and exploration probability $\epsilon=1$.
\For {$ i = 1: \#episodes $}
	\State Initialize game and get start state description $s$
	\While { \textbf{not} terminal}
    		\State $\phi_s = f_{\theta}(s)$
    		\State With probability $\epsilon$, sample a random action $a$, otherwise choose $\text{argmax}_{a} u_{\alpha}(\phi_{s}, a) \cdot \textbf{w}$
	        \State Execute $a$ and obtain next state $s'$ and reward $R(s')$ from environment
            \State Store transition $(s,a,R(s'),s')$ in $\mathcal{D}$
            \State Randomly sample mini-batches from $\mathcal{D}$
            \State Perform gradient descent on the loss $L^r(\mathbf{w}, \theta) + L^a(\tilde{\theta}, \theta)$ with respect to $\mathbf{w}$, $\theta$ and $\tilde{\theta}$. 
            \State Fix ($\theta,\tilde{\theta}, \mathbf{w})$ and perform gradient descent on $L^m(\alpha,\theta)$ with respect to $\alpha$.
            \State $s \leftarrow s'$
	    \EndWhile
	\State Anneal exploration variable $\epsilon$
\EndFor
\end{algorithmic}
\end{algorithm}

\section{Automatic Subgoal Extraction}
\label{sec:extractor}
Learning policies given sparse or delayed rewards is a significant challenge for current reinforcement learning algorithms. This is mainly due to inefficient exploration schemes such as $\epsilon-$greedy. Existing methods like Boltzmann exploration and Thomson sampling \cite{stadie2015incentivizing, osband2016deep}
offer significant improvements over $\epsilon$-greedy, but are limited due to the underlying models
functioning at the level of basic actions. Hierarchical reinforcement learning algorithms \cite{barto2003recent} such as the \textit{options} framework \cite{szepesvari2014universal,sutton1999between} provide a flexible framework to create temporal abstractions, which will enable exploration at different time-scales. The agent will learn options to reach the subgoals which can be used for intrinsic motivation. In the
context of hierarchical RL, \cite{goel2003subgoal} discuss a framework for subgoal extraction
using the structural aspects of a learned policy model. Inspired by previous work in subgoal discovery from state trajectories \cite{csimcsek2005identifying} and the tabular SR \cite{stachenfeld2014design}, we use the learned SR to generate plausible subgoal candidates. 

Given a random policy $\pi_r$ ($\epsilon=1$), we train the DSR until convergence and collect the SR for a large number of states $\mathcal{T}=\{m_{s_1, a_1},m_{s_2, a_2}, ..., m_{s_n, a_n} \}$. Following \cite{csimcsek2005identifying, shi2000normalized}, we generate an affinity matrix $W$ given $\mathcal{T}$, by applying a radial basis function (with Euclidean distance metric) for each pairwise entry $(m_{s_i,a_i}, m_{s_j,a_j})$ in $\mathcal{T}$ (to generate $w_{ij}$). Let $D$ be a diagonal matrix with $D(i,i)=\sum_j w_{ij}$. Then as per \cite{shi2000normalized}, the second largest eigenvalue of the matrix $D^{-1}(D-W)$ gives an approximation of the minimum normalized cut value of the partition of $\mathcal{T}$. The states that lie on the end-points of the cut are plausible subgoal candidates, as they provide a path between a community of state groups. Given randomly sampled $\mathcal{T}$ from $\pi_r$, we can collect statistics of how many times a particular state lies along the cut. We pick the top-\textit{k} states as the subgoals. Our experiments indicate that it is possible to extract useful subgoals from the DSR.

\section{Experiments}
In this section, we demonstrate the properties of our approach on MazeBase \cite{sukhbaatar2015mazebase}, a grid-world environment, and the Doom game engine \cite{kempka2016vizdoom}. In both environments, observations are presented as raw pixels to the agent. In the first experiment we show that our approach is comparable to DQN in two goal-reaching tasks. Next, we investigate the effect of modifying the distal reward on the initial Q-value. Finally, using normalized-cuts, we identify subgoals given the successor representations in the two environments.

\subsection{Goal-directed Behavior}
\label{sec:goal_exp}
\begin{figure}
    \centering
    \includegraphics[trim={0 3mm 0 -1mm}, scale=0.45]{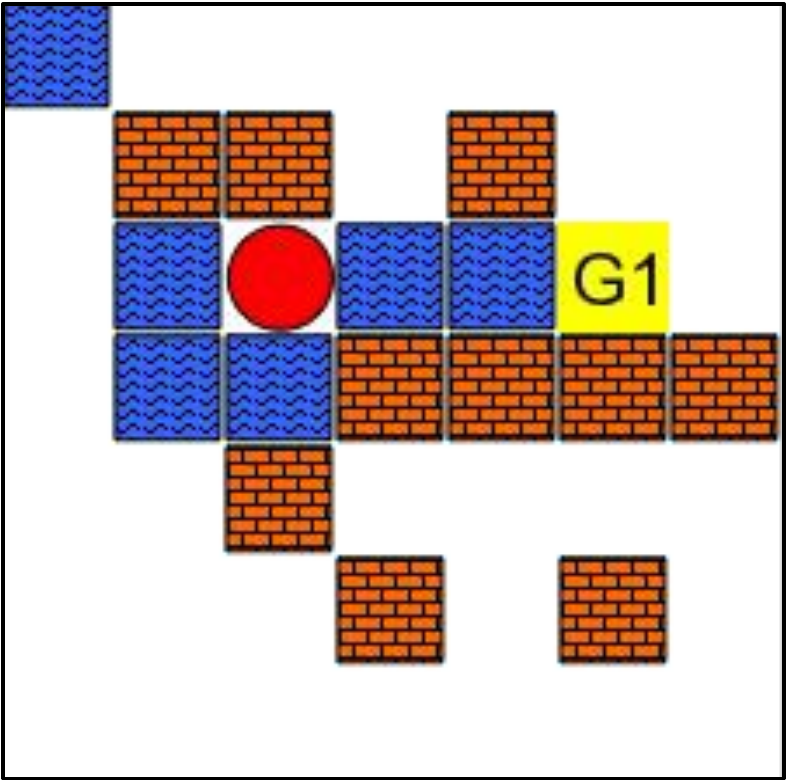}
    ~
     \centering
    \includegraphics[scale=0.15]{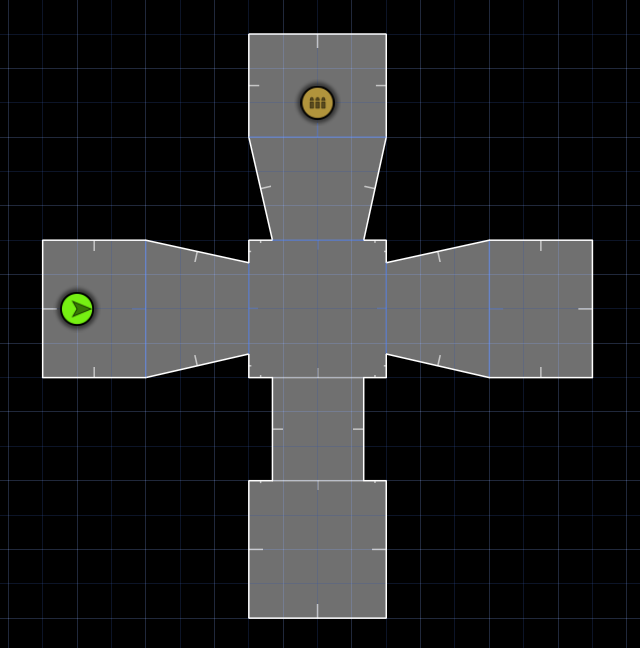}
    ~
     \centering
     \includegraphics[scale=0.3]{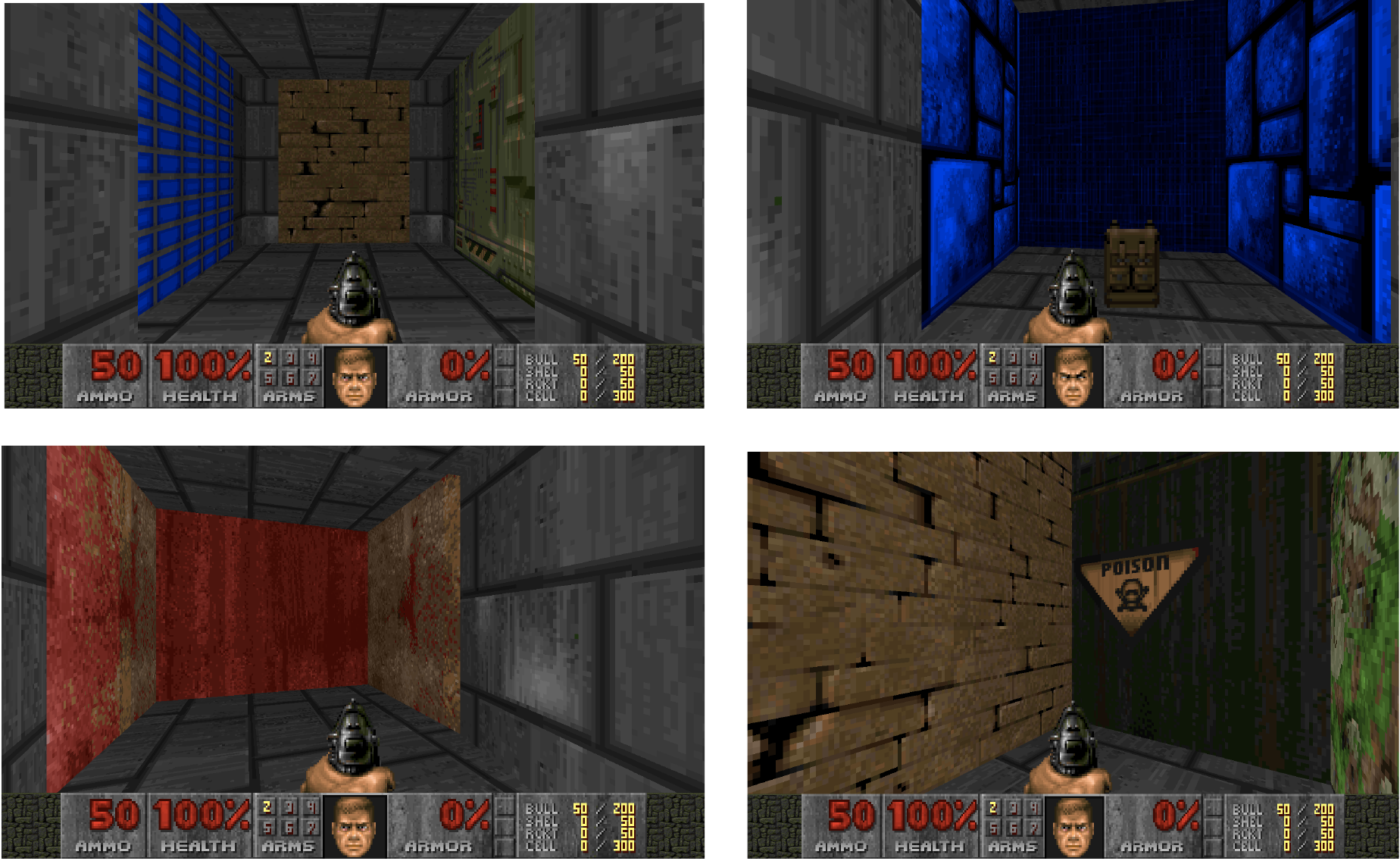}
      \caption{\textbf{Environments}: \textbf{(left)} MazeBase~\cite{sukhbaatar2015mazebase} map where the agent starts at an arbitrary location and needs to get to the goal state. The agent gets a penalty of -0.5 per-step, -1 to step on the water-block (blue) and +1 for reaching the goal state. The model observes raw pixel images during learning.  \textbf{(center)} A \textit{Doom} map using the VizDoom engine \cite{kempka2016vizdoom} where the agent starts in a room and has to get to another room to collect ammo (per-step penalty = -0.01, reward for reaching goal = +1). \textbf{(right)} Sample screen-shots of the agent exploring the 3D maze.}
          \label{fig:mazemap}
\end{figure}

\begin{figure*}
    \centering
    \includegraphics[width=0.47\textwidth]{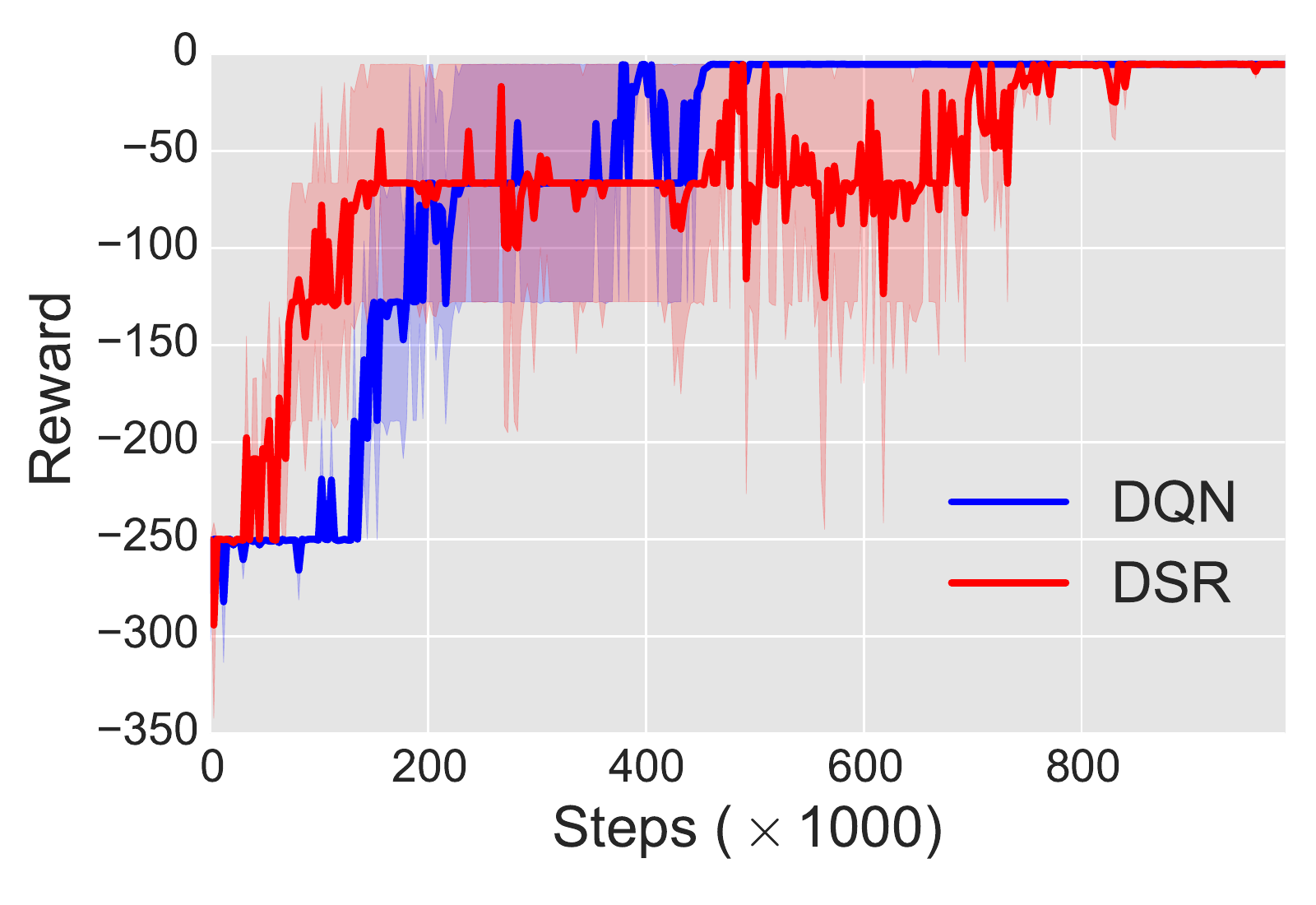}
    ~
     \centering
    \includegraphics[width=0.47\textwidth]{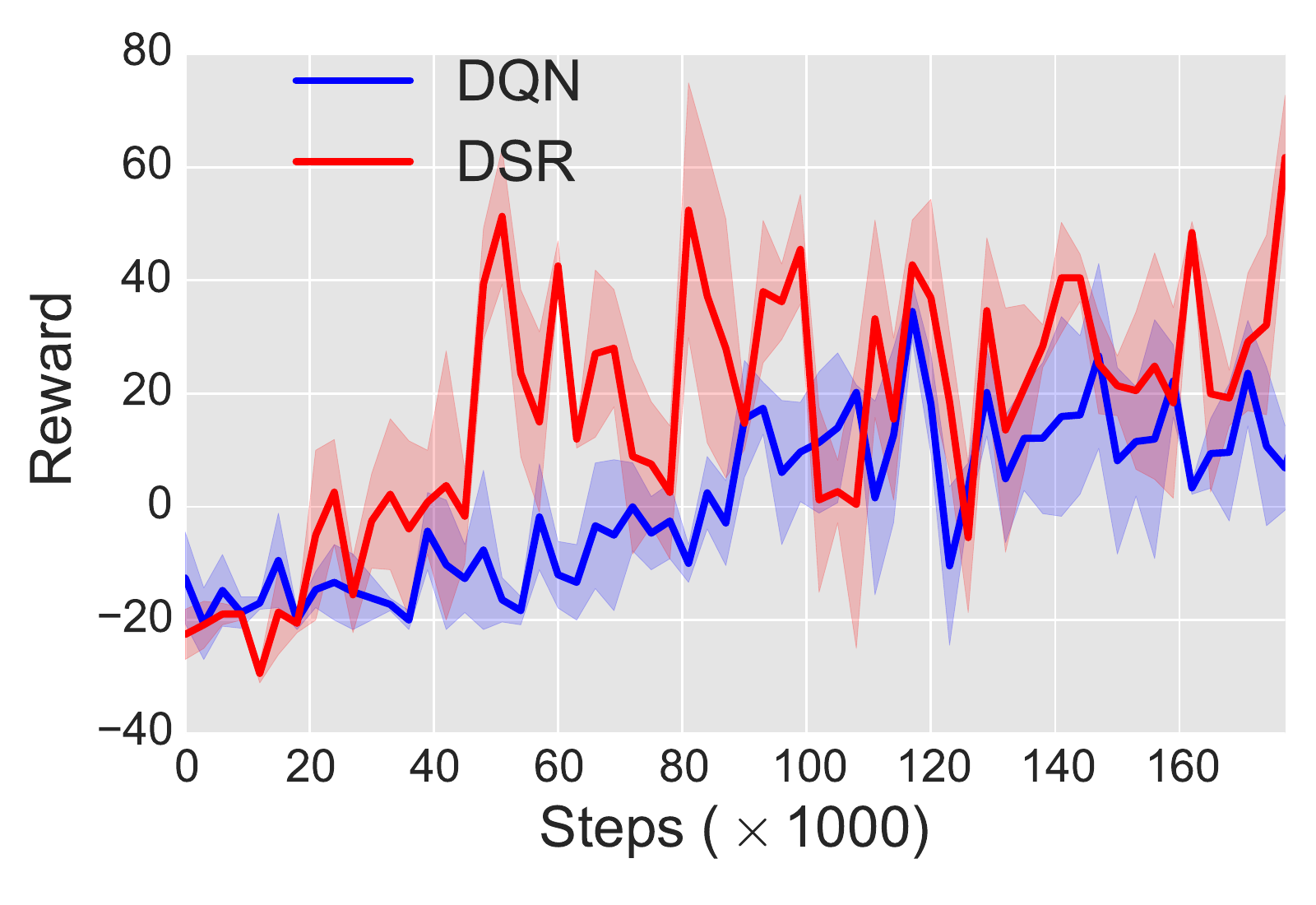}
      \caption{Average trajectory of the reward \textbf{(left)} over 100k steps for the grid-world maze. \textbf{(right)} over 180k steps for the Doom map over multiple runs.}
          \label{fig:mazeresults}
\end{figure*}

\paragraph{Solving a maze in MazeBase} We learn the optimal policy in the maze shown in Figure \ref{fig:mazemap} using the DSR and compare its performance to the DQN \cite{mnih2015human}. The cost of living or moving over water blocks is -0.5 and the reward value is 1. For this experiment, we set the discount rate to 0.99 and the learning rate to $2.5 \cdot 10^{-4}$. We anneal the $\epsilon$ from 1 to 0.1 over 20k steps; furthermore, for training the reward branch, we anneal the number of samples that we use, from 4000 to 1 by a factor of 0.5 after each training episode. For all experiments, we prioritize the reward training by keeping a database of non-zero rewards and sampling randomly from the replay buffer with a 0.8 probability and 0.2 from the database.  Figure \ref{fig:mazeresults} shows the average trajectory (over 5 runs) of the rewards obtained over 100k episodes. As the plot suggests, DSR performs on par with DQN. 

\paragraph{Finding a goal in a 3D environment} We created a map with 4 rooms using the ViZDoom platform~\cite{kempka2016vizdoom}. The map is shown in Figure \ref{fig:mazemap}. We share the same network architecture as in the case of MazeBase. The agent is spawned inside a room, and can explore any of the other three rooms. The agent gets a per-step penalty of -0.01 and a positive reward of 1.0 after collecting an item from one of the room (highlighted in \textit{red} in Figure\ref{fig:mazemap}). As shown in Figure\ref{fig:mazeresults}, the agent is able to successfully navigate the environment to obtain the reward, and is competitive with DQN.


\begin{figure}[h]
    \centering
    \hspace{10mm}
    \includegraphics[width=0.7\textwidth]{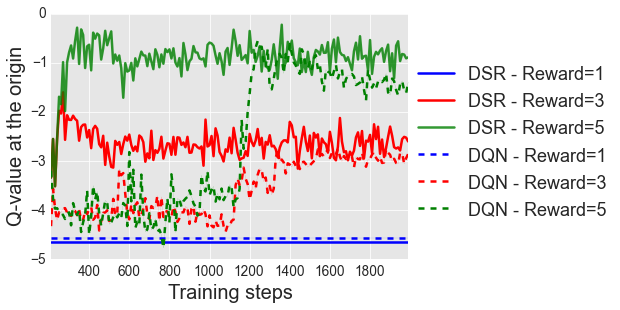}

      \caption{\textbf{Changing the value of the distal reward:} We train the model to learn the optimal policy on the maze shown in Figure \ref{fig:mazemap}. After convergence, we change the value of the distal reward and update the Q-value for the optimal action at the origin (bottom-left corner of the maze). In order for the value function to converge again, the model only needs to update the linear weights $\mathbf{w}$ given the new external rewards. }
          \label{fig:distal}
\end{figure}
\subsection{Value function sensitivity to distal reward changes}
The decomposition of value function into SR and immediate reward prediction allows DSR to rapidly adapt to changes in the reward function. In order to probe this, we performed experiments to measure the adaptability of the value function to distal reward changes. Given the grid-world map in Figure\ref{fig:mazemap}, we can train the agent to solve the goal specified in the map as highlighted in section \ref{sec:goal_exp}. Without changing the goal location, we can change the reward scalar value upon reaching the goal from 1.0 to 3.0. Our hypothesis is that due to the SR-based value decomposition, our value estimate will converge to this change by just updating the reward weights $\mathbf{w}$ (SR remains same). As shown in Figure~\ref{fig:distal}, we confirm that the DSR is able to quickly adapt to the new value function by just updating $\mathbf{w}$.

\subsection{Extracting subgoals from the DSR}
Following section \ref{sec:extractor}, we can also extract subgoals from the SR. We collect $\mathcal{T}$ by running a random policy on both MazeBase and VizDoom. During learning, we only update SR ($u_{\alpha}$) and the reconstruction branch ($g_{\tilde{\theta}}$), as the immediate reward at any state is zero (due to random policy). 

\begin{figure}[h!]
    \centering
    (a) \includegraphics[width=0.35\textwidth]{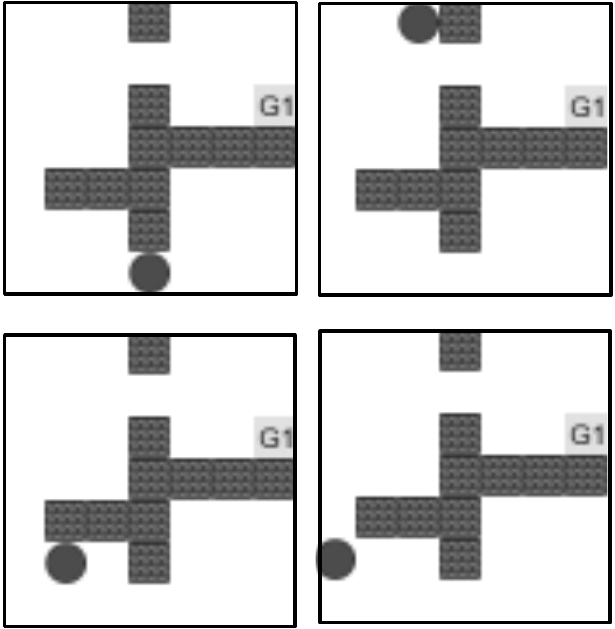}
    ~
     \centering
    (b)\includegraphics[width=0.35\textwidth]{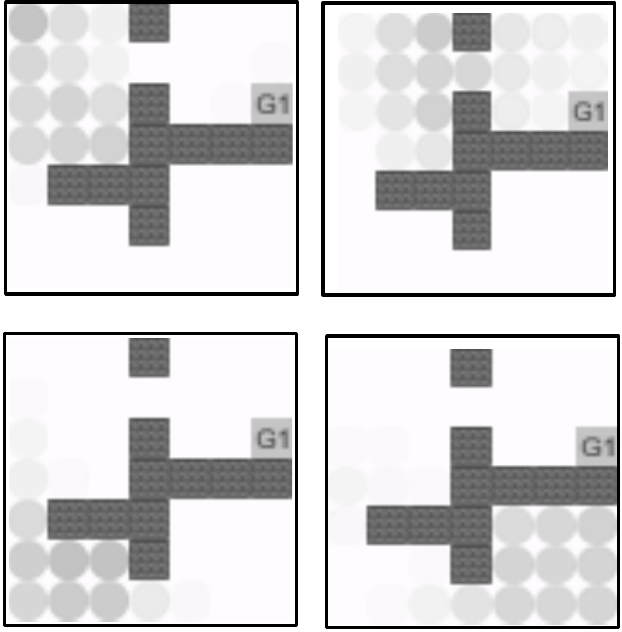}
      \caption{\textbf{Subgoal extraction on grid-world:} Given a random policy, we train DSR until convergence and collect a large number of sample transitions and their corresponding successor representations as described in section~\ref{sec:extractor}. We apply a normalized cut-based algorithm on the SRs to obtain a partition of the environment as well as the bottleneck states (which correspond to goals)  \textbf{(a)} Subgoals are states which separate different partitions of the environments under the normalized-cut algorithm. Our approach is able to find reasonable subgoal candidates. \textbf{(b)} Partitions of the environment reflect latent structure in the environment.}
          \label{fig:mb_subgoals}
\end{figure}

\begin{figure*}[h!]
    \centering
    \includegraphics[width=0.6\textwidth]{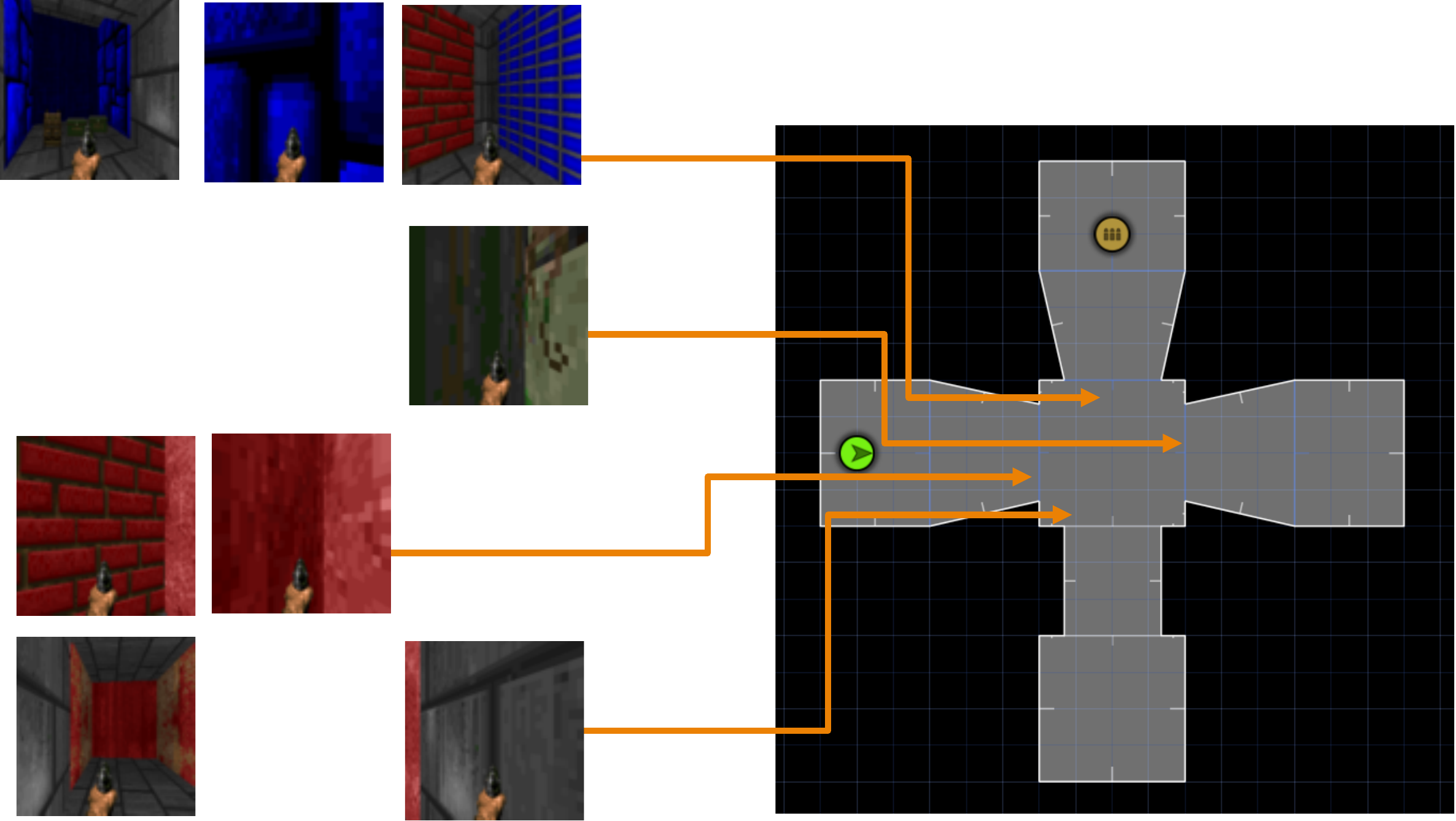}
      \caption{\textbf{Subgoal extraction on the Doom map} The subgoals are extracted using the normalized cut-based algorithm on the SR. The SR samples are collected based on a random policy. The subgoals mostly correspond to the rooms' entrances in the common area between the rooms. Due to random policy, we sometimes observe high variance in the subgoal quality. Future work should address robust statistical techniques to obtain subgoals, as well as non-parametric approaches to obtaining flexible number of subgoals.}
          \label{fig:doom_subgoals}
\end{figure*}

As shown in Figures \ref{fig:mb_subgoals} and \ref{fig:doom_subgoals}, our subgoal extraction scheme is able to capture useful subgoals and clusters the environment into reasonable segments. Such a scheme can be ran periodically within a hierarchical reinforcement learning framework to aid exploration. One inherent limitation of this approach is that due to the random policy, the subgoal candidates are often quite noisy. Future work should address this limitation and provide statistically robust ways to extract plausible candidates. Additionally, the subgoal extraction algorithm should be non-parametric to handle flexible number of subgoals. 

\section{Conclusion}
We presented the DSR, a novel deep reinforcement learning framework to learn goal-directed behavior given raw sensory observations. The DSR estimates the value function by taking the inner product between the SR and immediate reward predictions. This factorization of the value function gives rise to several appealing properties over existing deep reinforcement learning methods---namely increased sensitivity of the value function to distal reward changes and the possibility of extracting subgoals from the SR under a random policy. 

For future work, we plan to combine the DSR with hierarchical reinforcement learning. Learning goal-directed behavior with sparse rewards is a fundamental challenge for existing reinforcement learning algorithms. The DSR can enable efficient exploration by periodically extracting subgoals, learning policies to satisfy these intrinsic goals (skills), and subsequently learning hierarchical policy over these subgoals in an options framework \cite{szepesvari2014universal, kulkarni2016hierarchical, schaul2015universal}. One of the major issues with the DSR is learning discriminative features. In order to scale up our approach to more expressive environments, it will be crucial to combine various deep generative and self-supervised models \cite{eslami2016attend, greff2015binding, rezende2016one, kulkarni2015deep, whitney2016understanding, gregor2015draw, huang2015efficient, noroozi2016unsupervised} with our approach. In addition to subgoals, using DSR for extracting other intrinsic motivation measures such as improvements to the predictive world model \cite{schmidhuber2010formal} or mutual information \cite{mohamed2015variational} is worth pursuing. 


\renewcommand*{\bibfont}{\small}
\bibliographystyle{plain}
\bibliography{deepSR}

\end{document}